\documentclass[lettersize,journal]{IEEEtran}
\usepackage[caption=false,font=normalsize,labelfont=sf,textfont=sf]{subfig}
\usepackage{stfloats}
\usepackage{cite}
\newcommand{\fref}[1]{Figure \ref{#1}}
\newcommand{\sref}[1]{Section \ref{#1}}
\newcommand{\tref}[1]{Table \ref{#1}}

\newcommand{\alref}[1]{Algorithm \ref{#1}}
\usepackage{graphicx}
\usepackage{amsmath}
\usepackage{amsthm}

\usepackage{amssymb,amsfonts}
\usepackage{algorithmicx}
\usepackage{array}
\usepackage{multirow}
\usepackage{mathrsfs}%
\usepackage{xcolor}%
\usepackage{textcomp}%
\usepackage{manyfoot}%
\usepackage{booktabs}%
\usepackage{algorithm}%
\usepackage{algpseudocode}%
\usepackage{listings}%
\usepackage{hyperref}
\usepackage{balance}
\usepackage{tabularx}
\usepackage{enumitem}
\setlist[enumerate]{label={(\arabic*)}}
\usepackage{soul}
\usepackage{array}
\usepackage{stfloats}
\usepackage{url}
\usepackage{verbatim}
\usepackage{epsfig}
\usepackage{color}
\usepackage{colortbl}
\usepackage{lineno}
\usepackage{graphics}
\usepackage{lmodern}
\graphicspath{{figures/}}
\usepackage{rotating}
\usepackage{float}
\usepackage{cases}
\usepackage{supertabular}
\usepackage{makecell}
\usepackage{url}

\begin{document}

\title{Large Language Model Assisted Automated Algorithm Generation and Evolution via Meta-black-box optimization}

\author{Xu~Yang,
        Rui~Wang,
        Kaiwen~Li,
        Wenhua~Li,
        and~Weixiong~Huang
\thanks{Xu Yang, Rui Wang, Kaiwen Li, Wenhua Li and Weixiong Huang are with the College of Systems Engineering, National University of Defense Technology, Changsha 410073, China (e-mail: yangxu616@nudt.edu.cn; rui\_wang@nudt.edu.cn; likaiwen@nudt.edu.cn, liwenhua@nudt.edu.cn, huangweixiong@nudt.edu.cn).}
\thanks{Manuscript received September 17, 2025 (\textit{Corresponding author: Rui Wang})}}
\markboth{Journal of \LaTeX\ Class Files,~Vol.~14, No.~8, August~2025}%
{Xu Yang \MakeLowercase{\textit{et al.}}: A Sample Article Using IEEEtran.cls for IEEE Journals}


\maketitle

\begin{abstract}
    Meta-black-box optimization has been significantly advanced through the use of large language models (LLMs), yet in fancy on constrained evolutionary optimization. In this work, AwesomeDE is proposed that leverages LLMs as the strategy of meta-optimizer to generate update rules for constrained evolutionary algorithm without human intervention. On the meanwhile, $RTO^2H$ framework is introduced for standardize prompt design of LLMs. The meta-optimizer is trained on a diverse set of constrained optimization problems. Key components, including prompt design and iterative refinement, are systematically analyzed to determine their impact on design quality. Experimental results demonstrate that the proposed approach outperforms existing methods in terms of computational efficiency and solution accuracy. Furthermore, AwesomeDE is shown to generalize well across distinct problem domains, suggesting its potential for broad applicability. This research contributes to the field by providing a scalable and data-driven methodology for automated constrained algorithm design, while also highlighting limitations and directions for future work.
\end{abstract}
  
\section{Introduction}
Traditional evolutionary algorithms (EAs) to constrained optimization problems (COPs) often rely on handcrafted evolutionary strategies \cite{runarsson2000stochastic,zuo2023process,ozkaya2023fitness,qiao2023self} and constraint handling techniques \cite{peng2023evolutionary,wang2024adaptive,rahimi2023review}, which may lack adaptability across diverse problem domains. Meta-black-box optimization (MetaBBO) \cite{yang2025metareview} has been transformed by recent advances in large language models (LLMs), particularly in the domain of EAs \cite{ZHONG2025103042,zhong2024geminide,NEURIPS2023_184c1e18,lange2024llm,liu2024large,zhong2024leveraging}. While LLMs have demonstrated remarkable capabilities, their application to constrained evolutionary optimization (CEO) remains underexplored \cite{wang2024large}. 

In this work, a data-driven framework is proposed that harnesses LLMs to automatically design update rules of constrained evolutionary algorithms (CEAs) via MetaBBO for COPs. The approach is distinguished by its elimination of human intervention in the design process, while maintaining rigorous adherence to constraint satisfaction. 

The proposed framework is systematically evaluated across multiple benchmark COPs, with particular attention paid to computational efficiency and solution accuracy. This research makes three primary contributions: (1) a generalizable LLM-based EA using EA-based MetaBBO for constrained evolutionary algorithm design that requires no human interactions, (2) Enhancing interpretability via update rule generation, and (3) empirical validation showing superior performance compared to existing hand-designed methods. The results suggest that LLMs can indeed learn principled update strategies that respect problem constraints while maintaining evolutionary algorithm effectiveness.

The remainder of this paper is organized as follows: Section~\ref{sec:related_work} discusses relevant literature, Section~\ref{sec:method} details the proposed framework, Section~\ref{sec:experiments} presents experimental results, and Section~\ref{sec:conclusion} concludes with broader implications and future directions.

\section{Related Work}
\label{sec:related_work}
Recent advancements in Large Language Models (LLMs) have catalyzed transformative applications across algorithmic design paradigms. This section contextualizes our work within the broader landscape of LLM-centric algorithmic innovation, with a focal emphasis on CEO.

\subsection{LLM-Driven Evolutionary Optimization}
The integration of LLMs into EAs has emerged as a paradigm-shifting approach for MetaBBO. Notable studies include the utilization of LLMs as black-box optimizers within EA frameworks. In single-objective optimization, LLM is used to assist hyper-heuristic optimization \cite{ZHONG2025103042,zhong2024geminide}, algorithm generation \cite{NEURIPS2023_184c1e18,zhong2024leveraging}, and solution manipulation \cite{liu2024large,lange2024llm}. In multi-objective optimization, LLM-augmented frameworks exhibit capability to generate Pareto-optimal solutions through decomposition-based strategies \cite{liu2025large}. However, existing works predominantly focus on unconstrained optimization landscapes, leaving a critical gap in addressing complex constraint satisfaction challenges inherent in real-world engineering problems.

\subsection{Constraint Evolutionary Optimization}
Traditional CEO methodologies rely on constraint handling techniques such as penalty functions, feasibility rule, multi-objective method and hybrid method \cite{wang2009constrained,peng2023evolutionary}. Recent efforts have explored machine learning integration for constraint modeling \cite{rana2020recent,popescu2022overview}, yet these approaches often lack adaptability and generalizability. The application of LLMs in this domain remains nascent, with preliminary studies demonstrating feasibility in speeding up the convergence of the evolutionary population\cite{wang2024large}. Nevertheless, challenges persist in ensuring constraint compliance during solution generation and maintaining diversity under constrained search spaces.

\subsection{LLM-Enhanced Constrained Evolutionary Optimization}
Emerging research has begun to explore LLM-driven frameworks for automated constraint handling in evolutionary optimization. For instance, the authors in \cite{wang2024large} finetune the LLM through tailored prompt engineering, integrating information concerning both objective values and constraint violations of solutions. By leveraging the refined LLM, it can be used as a search operator to generate superior-quality solutions. However, it is lack of interpretability.

\section{Method}
\label{sec:method}
This section provides a detailed explanation of the proposed LLM-based meta-optimizer to design a constrained evolutionary algorithm (llmEA) for COPs. First, the design motivation is analyzed, followed by a detailed description of the algorithmic framework. The meta-training mechanism through MetaBBO is then formalized, with particular emphasis on the specialized prompt engineering strategy.

\subsection{Motivation}
The challenge faced by COPs lies in effectively finding optimal feasible solution, which is difficult due to the presence of various kinds of constraints. It may lead to uncontinuous and/or small feasible regions. While previous research has proposed various search strategies or constraint handling techniques, the emergence of LLMs offers new possibilities due to their impressive capabilities in reasoning and predicting. Integrated into MetaBBO, this adaptability enables LLMs to design algorithms for COPs without human interactions, showcasing exceptional generalization performance. Specifically, LLM will output an optimal update rule for the given COP considering the decision variable information, objective information and constraints information, which enhances much interpretability.

\subsection{Algorithmic Framework}
The pseudocode of llmEA is presented in \alref{algLLM4CO}, where the input LLM-based meta-optimizer is trained-done via MetaBBO (introduced in \sref{secMetaBBO}). 

\begin{algorithm}
    \caption{General framework of the proposed llmEA}
    \begin{algorithmic}[1]
    \Require population size $N$, maximum function evaluations $maxFE$, trained LLM-based meta-optimizer 
    \Ensure the final population $P$
    \State $\mathcal{F} \gets$ output an update rule by $MO$
    \State $P \gets$ initialize the population consisting of $N$ solutions
    \State $FE \gets 0$
    \While {$FE <= maxFE$}
    \State $O \gets $ generate $N$ offspring by the update rule $\mathcal{F}$ applying on $P$
    \State $U \gets P \cup O$
    \State $P \gets$ select $N$ solutions from $U$ by the environment selection
    \State $FE = FE + N$
    \EndWhile
    \State \Return the final population $P$
    \end{algorithmic}
    \label{algLLM4CO}
  \end{algorithm}
  
First, an update rule $\mathcal{F}$ is derived from the trained LLM to guide the optimization process. The population is then initialized with $N$ randomly generated solutions. During each iteration, $N$ offspring solutions are generated by applying the learned update rule $\mathcal{F}$ to the current population. The combined solution set $U$, formed by merging the parent and offspring populations, undergoes environmental selection to maintain population size constraints. This selection mechanism preserves high-quality solutions while maintaining diversity. The function evaluation counter is incremented by $N$ after each iteration, reflecting the parallel evaluation of solutions. The process continues until the maximum number of function evaluations is reached, ensuring efficient resource utilization while exploring the solution space.

\subsection{Meta-training LLM-based meta-optimizer}
\label{secMetaBBO}
\begin{figure*}
    \centering
    \includegraphics[width = 0.95\textwidth]{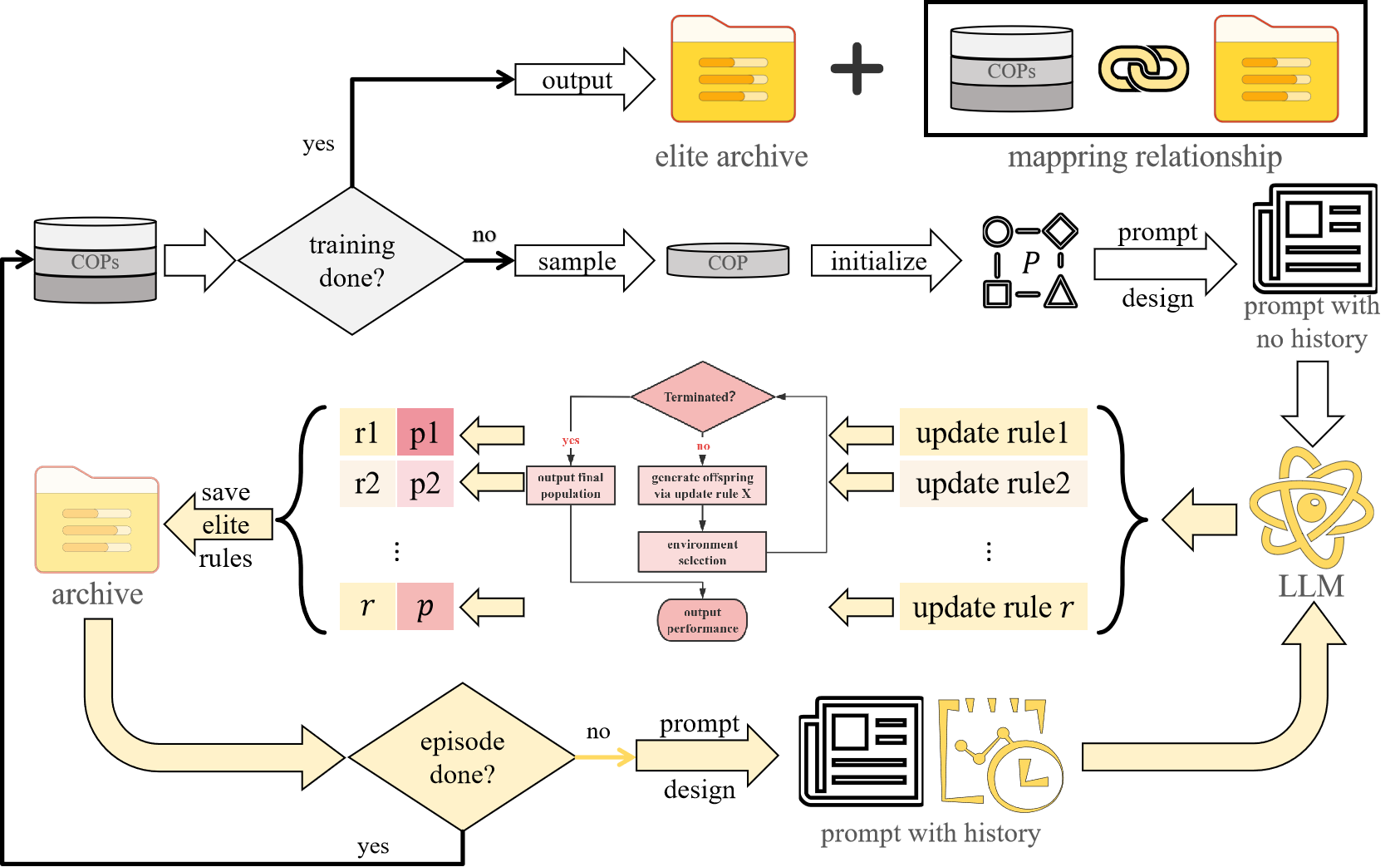}
    \caption{Illustrating the integration of a LLM as meta-optimizer to design constrained evolutionary algorithms via MetaBBO. The yellow hollow arrow points to the outer loop, and the pink flowchart in the middle of the picture represents the inner loop.}
    \label{figIllustration}
  \end{figure*}

MetaBBO leverages a bi-level optimization framework to discover or refine black-box optimization algorithms including CEAs via meta-learning. The LLM is employed as a meta-optimizer to systematically design CEAs by evolving update rules based on performance feedback. For a detailed illustration of how LLM designs constrained evolutionary algorithm via MetaBBO, refer to \fref{figIllustration}.  As depicted in \fref{figIllustration}, the framework comprises two interdependent loops: an outer meta-optimization loop (yellow arrows) and an inner algorithm execution loop (pink flowchart).

In the outer loop, historical optimization information and elite update rules are archived to guide the LLM's update rule generation. At each meta-training iteration, the LLM produces candidate update rules conditioned on the archived information. These rules are made up of CEAs and then deployed in the inner loop to solving the sampled COP. The inner loop evaluates generated rules by executing CEAs across training problem instances, with performance metrics fed back to the outer loop. Historical trajectories, including CEAs'performance, are systematically recorded to enhance rule evolution.

\subsection{Prompt design}
In the MetaBBO, the LLM serves as a meta-optimizer for generating novel update rules. It is crucial to recognize that LLM relies on prompts to perform tasks. Prompt design tailored for automatically designing CEAs using LLM is craft to address COPs, with a focus on improving the solution quality and feasibility. These prompts typically consists of five main components:
\begin{itemize}
    \item Role Definition: this part defines the LLM role, briefly describes who the LLM acts and what to do in the current episode.
    \item Task Description: this part provides the detailed task description covering information about decision variables, constraint violations and objective values of the initial population.
    \item Operating Requirement: this part tells the specific steps employed by LLM to generate update rules and requirements.
    \item History Feedback: this part provides the history records including generated update rules and their experimental performance of solving the sampled COP.
    \item Output Format: this part defines the output format of LLM to facilitate the extraction and process the results.
\end{itemize}
 
\begin{figure*}
    \centering
    \includegraphics[width = 0.95\textwidth]{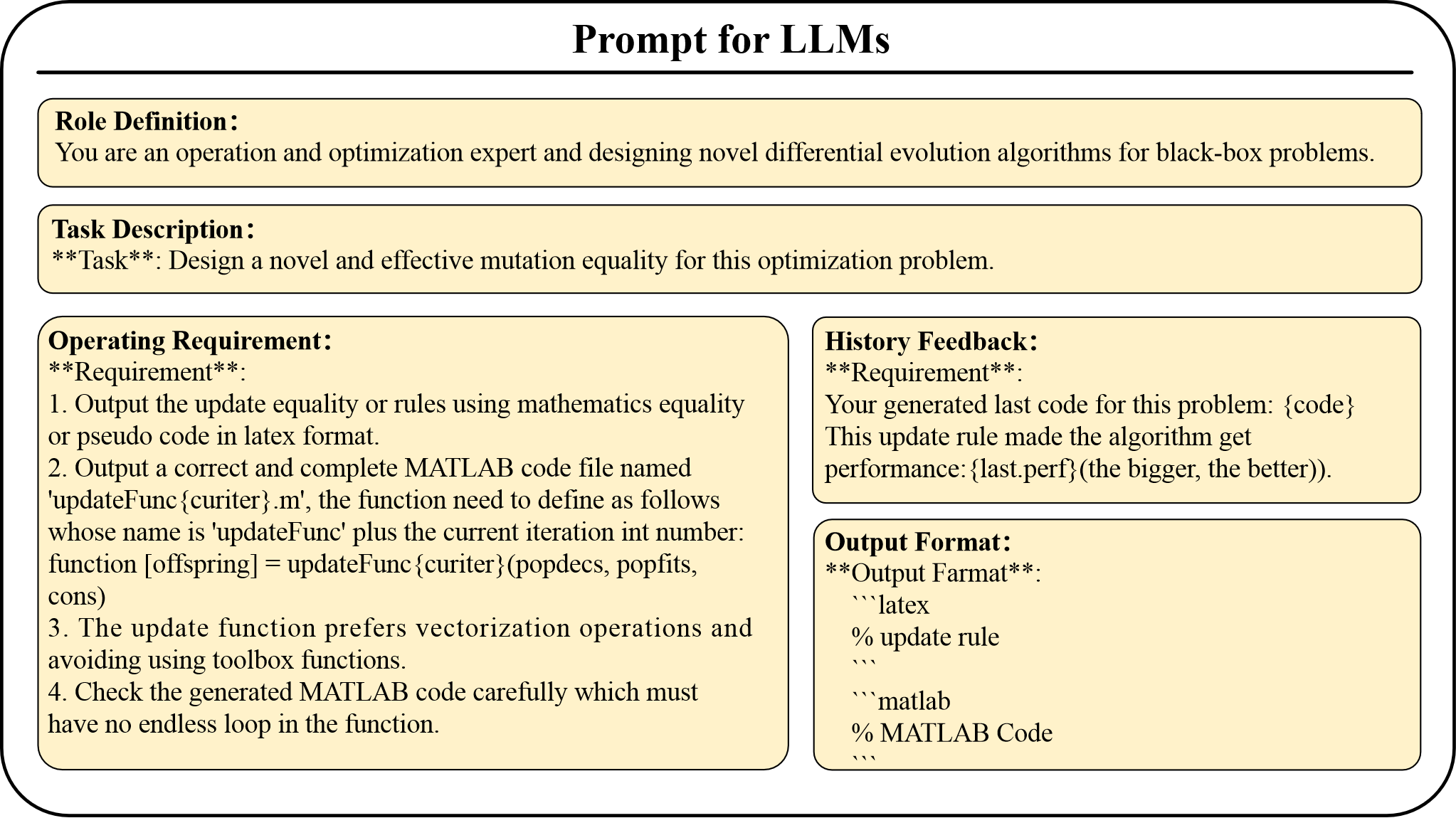}
    \caption{An example of a prompt to guide LLM for designing a CEA specifically the update rule.}
    \label{figPrompt}
  \end{figure*}

An example prompt is illustrated as \fref{figPrompt}. 

  This structured knowledge enables the LLM to propose update rules that balance objective optimization with systematic constraint satisfaction.

\section{Experiments and Results}
\label{sec:experiments}
\subsection{Experimental setting}
The CEC2010 benchmark suite for constrained real-parameter optimization \cite{mallipeddi2010problem} is adopted to evaluate the proposed method. This standardized test set comprises 18 scalable problems with diverse characteristics, including separable/non-separable objectives, inequality/equality constraints, and rotated constraint geometries. Each problem is formulated as:

\begin{equation}
    \text{Minimize } f(\mathbf{x}), \quad \mathbf{x} \in S \subseteq \mathbb{R}^D
\end{equation}
subject to:
\begin{equation}
    g_i(\mathbf{x}) \leq 0 \ (1 \leq i \leq p), \quad |h_j(\mathbf{x})| \leq \epsilon \ (p+1 \leq j \leq m)
\end{equation}
where $\epsilon=10^{-4}$ denotes the equality tolerance. The search space dimensionality $D$ is configured as 10 and 30, with maximum function evaluations (MaxFE) set to $2\times10^5$ and $6\times10^5$, respectively. Key problem characteristics include: (1) Rotated constraints that eliminate coordinate system bias, (2) Feasibility ratios ranging from 0 to 1, and (3) Hybrid constraint types per problem. 

Furthermore, classical algorithm differential evolution algorithm (DE) \cite{storn1997differential}, manually-improved algorithms IMODE \cite{sallam2020improved} and SHADE \cite{tanabe2013success} are selected as comparative algorithms to assess the effectiveness of llmEA. To ensure fair comparisons, the parameters of the comparative algorithms are set according to the original papers. In llmEA, Deepseek R1 is leveraged as the LLM which generates 5 update rules in a loop. 

All experiments are conducted over 31 independent runs to ensure statistical significance. Performance metrics include best/median/worst objective values and feasibility rates. Algorithm complexity is quantified through normalized computation times $T_1$ (pure evaluation cost) and $T_2$ (full optimization overhead). The evaluation protocol strictly adheres to CEC2010 specifications, where constraint function evaluations are counted toward MaxFE. All experiments are conducted on the PlatMetaX \cite{yang2025platmetax}.

\subsection{Results on CEC2010 benchmark suite}
The comparative performance of llmEA against competitors is systematically presented through \tref{tabMinVResCEC2010} on average minimum objective values $v_{avg}$ and standard deviations $v_{std}$. The best results are highlighted with gray background. Significant performance differences between llmEA and competitors are statistically validated through pairwise comparisons, as indicated by the "+", "-", and "=" symbols. If an algorithm fails to find any feasible solution, the corresponding result is marked as "NaN". 

It is observed that llmEA achieves the best performance on six benchmark functions (C02, C09, C14-C18), demonstrating superior exploration capabilities in complex search landscapes. Particularly noteworthy are its solutions for C16-C18, where the average minimum values are reduced by 55.1\%-96.5\% compared to the second-best results. The algorithm's effectiveness in handling high-dimensional constraints is evidenced by its unique success in generating feasible solutions for C09, C14, and C15, where all competitors failed.

LSHADE and MadDE exhibit competitive performance on specific functions (C01, C13 and C05, C11 respectively), suggesting complementary strengths in different problem types. However, their limited success rates on 6 and 7 functions respectively highlight reliability issues in constrained optimization scenarios. Traditional DE shows consistent but suboptimal performance, being outperformed by llmEA in 5 of 12 comparable instances.

The prevalence of NaN entries (63.6\% of all results) underscores the benchmark's difficulty. llmEA demonstrates improved robustness with feasible solutions obtained for 66.7\% of test functions, compared to 33.3\%-50\% for other algorithms. The standard deviations in llmEA's results (e.g., 8.56e-1 for C02 and 2.50e-1 for C16) suggest stable convergence patterns, particularly when compared to the wider performance fluctuations observed in GA and J21.

These findings collectively demonstrate llmEA's advancements in balancing exploration-exploitation tradeoffs and constraint handling. The algorithm's ability to maintain solution feasibility while achieving competitive objective values positions it as a promising approach for real-world optimization problems with complex constraints.

\begin{table}[htbp] \tiny
    \centering
    \caption{Per-instance optimization results $v_{avg}(v_{std})$ of each algorithm on CEC2010 with $D=10$, where the symbols "-","+", and "=" separately indicate whether the given algorithm performs significantly worse, significantly better, or equal to llmEA.}
    \begin{tabular}{cccccccc}
        \toprule
        \textbf{$P$} & \textbf{GA} & \textbf{DE} & \textbf{J21} & \textbf{LSHADE} & \textbf{MadDE} & \textbf{NLC} & \textbf{llmEA} \\
        \midrule
        C01   &  \makecell{-6.7626e-1 \\(5.97e-2) +} &  \makecell{-4.7769e-1 \\(2.91e-2) -} &  \makecell{-7.1875e-1 \\(1.87e-2) +} &  \makecell{\cellcolor[rgb]{ .816,  .808,  .808}-7.3561e-1 \\\cellcolor[rgb]{ .816,  .808,  .808}(1.67e-2) +} &  \makecell{-7.3051e-1 \\(2.25e-2) +} &  \makecell{-6.4699e-1 \\(3.11e-2) =} &  \makecell{-6.1086e-1 \\(9.87e-2)} \\
        C02   &  \makecell{3.5793e+0 \\(1.04e+0) -} &  \makecell{4.3222e+0 \\(8.35e-1) -} &  \makecell{4.0853e+0 \\(0.00e+0) =} &  \makecell{3.1366e+0 \\(1.25e+0) -} &  \makecell{3.7136e+0 \\(1.15e+0) -} &  \makecell{3.9181e+0 \\(1.12e+0) -} &  \makecell{\cellcolor[rgb]{ .816,  .808,  .808}-1.1205e+0 \\\cellcolor[rgb]{ .816,  .808,  .808}(8.56e-1)} \\
        C03   &  \makecell{NaN \\(NaN)} &  \makecell{NaN \\(NaN)} &  \makecell{NaN \\(NaN)} &  \makecell{\cellcolor[rgb]{ .816,  .808,  .808}8.7486e+5 \\\cellcolor[rgb]{ .816,  .808,  .808}(8.46e+5)} &  \makecell{1.0520e+6 \\(1.26e+6)} &  \makecell{NaN \\(NaN)} &  \makecell{NaN \\(NaN)} \\
        C04   &  \makecell{NaN \\(NaN)} &  \makecell{NaN \\(NaN)} &  \makecell{NaN \\(NaN)} &  \makecell{1.5484e+0 \\(4.86e+0)} &  \makecell{1.4501e-1 \\(3.04e-1)} &  \makecell{NaN \\(NaN)} &  \makecell{NaN \\(NaN)} \\
        C05   &  \makecell{NaN \\(NaN)} &  \makecell{NaN \\(NaN)} &  \makecell{NaN \\(NaN)} &  \makecell{NaN \\(NaN)} &  \makecell{\cellcolor[rgb]{ .816,  .808,  .808}2.1359e+2 \\\cellcolor[rgb]{ .816,  .808,  .808}(0.00e+0)} &  \makecell{NaN \\(NaN)} &  \makecell{NaN \\(NaN)} \\
        C06   &  \makecell{\cellcolor[rgb]{ .816,  .808,  .808}2.9308e+2 \\\cellcolor[rgb]{ .816,  .808,  .808}(0.00e+0)} &  \makecell{NaN \\(NaN)} &  \makecell{NaN \\(NaN)} &  \makecell{NaN \\(NaN)} &  \makecell{4.9115e+2 \\(3.15e+1)} &  \makecell{NaN \\(NaN)} &  \makecell{NaN \\(NaN)} \\
        C07   &  \makecell{1.0167e+2 \\(2.33e+2) +} &  \makecell{1.8394e+7 \\(2.94e+7) +} &  \makecell{2.4823e+3 \\(5.61e+3) +} &  \makecell{\cellcolor[rgb]{ .816,  .808,  .808}8.4558e-1 \\\cellcolor[rgb]{ .816,  .808,  .808}(1.62e+0) +} &  \makecell{1.4506e+0 \\(1.48e+0) +} &  \makecell{2.1461e+6 \\(8.73e+6) +} &  \makecell{1.0211e+9 \\(1.01e+9)} \\
        C08   &  \makecell{NaN \\(NaN)} &  \makecell{NaN \\(NaN)} &  \makecell{NaN \\(NaN)} &  \makecell{NaN \\(NaN)} &  \makecell{NaN \\(NaN)} &  \makecell{NaN \\(NaN)} &  \makecell{NaN \\(NaN)} \\
        C09   &  \makecell{NaN \\(NaN)} &  \makecell{NaN \\(NaN)} &  \makecell{NaN \\(NaN)} &  \makecell{NaN \\(NaN)} &  \makecell{NaN \\(NaN)} &  \makecell{NaN \\(NaN)} &  \makecell{\cellcolor[rgb]{ .816,  .808,  .808}3.0537e+7 \\\cellcolor[rgb]{ .816,  .808,  .808}(5.08e+7)} \\
        C10   &  \makecell{NaN \\(NaN)} &  \makecell{NaN \\(NaN)} &  \makecell{NaN \\(NaN)} &  \makecell{NaN \\(NaN)} &  \makecell{NaN \\(NaN)} &  \makecell{NaN \\(NaN)} &  \makecell{NaN \\(NaN)} \\
        C11   &  \makecell{NaN \\(NaN)} &  \makecell{NaN \\(NaN)} &  \makecell{NaN \\(NaN)} &  \makecell{\cellcolor[rgb]{ .816,  .808,  .808}-2.7296e-4 \\\cellcolor[rgb]{ .816,  .808,  .808}(8.59e-4)} &  \makecell{-1.7445e-4 \\(0.00e+0)} &  \makecell{NaN \\(NaN)} &  \makecell{NaN \\(NaN)} \\
        C12   &  \makecell{NaN \\(NaN)} &  \makecell{NaN \\(NaN)} &  \makecell{NaN \\(NaN)} &  \makecell{2.4233e+0 \\(1.37e+1)} &  \makecell{\cellcolor[rgb]{ .816,  .808,  .808}6.4210e-1 \\\cellcolor[rgb]{ .816,  .808,  .808}(3.02e+1)} &  \makecell{1.2563e+1 \\(0.00e+0)} &  \makecell{NaN \\(NaN)} \\
        C13   &  \makecell{-6.3160e+1 \\(2.66e+0) +} &  \makecell{-4.9853e+1 \\(1.96e+0) -} &  \makecell{-6.0696e+1 \\(2.92e+0) +} &  \makecell{\cellcolor[rgb]{ .816,  .808,  .808}-6.6016e+1 \\\cellcolor[rgb]{ .816,  .808,  .808}(1.37e+0) +} &  \makecell{-6.6023e+1 \\(2.17e+0) +} &  \makecell{-6.5557e+1 \\(9.52e-1) +} &  \makecell{-5.5510e+1 \\(4.01e-1)} \\
        C14   &  \makecell{NaN \\(NaN)} &  \makecell{NaN \\(NaN)} &  \makecell{1.0657e+14 \\(4.98e+13) -} &  \makecell{NaN \\(NaN)} &  \makecell{1.4088e+14 \\(0.00e+0) =} &  \makecell{NaN \\(NaN)} &  \makecell{\cellcolor[rgb]{ .816,  .808,  .808}1.1694e+7 \\\cellcolor[rgb]{ .816,  .808,  .808}(4.11e+7)} \\
        C15   &  \makecell{NaN \\(NaN)} &  \makecell{NaN \\(NaN)} &  \makecell{NaN \\(NaN)} &  \makecell{NaN \\(NaN)} &  \makecell{NaN \\(NaN)} &  \makecell{NaN \\(NaN)} &  \makecell{\cellcolor[rgb]{ .816,  .808,  .808}1.6451e+8 \\\cellcolor[rgb]{ .816,  .808,  .808}(1.89e+8)} \\
        C16   &  \makecell{1.0522e+0 \\(2.02e-2) -} &  \makecell{NaN \\(NaN)} &  \makecell{1.0853e+0 \\(3.25e-2) -} &  \makecell{1.0110e+0 \\(5.34e-2) -} &  \makecell{1.0184e+0 \\(4.00e-2) -} &  \makecell{1.0823e+0 \\(4.45e-2) -} &  \makecell{\cellcolor[rgb]{ .816,  .808,  .808}4.5437e-1 \\\cellcolor[rgb]{ .816,  .808,  .808}(2.50e-1)} \\
        C17   &  \makecell{5.5464e+2 \\(2.81e+2) -} &  \makecell{1.2311e+3 \\(6.15e+2) -} &  \makecell{9.0826e+2 \\(0.00e+0) =} &  \makecell{3.3998e+2 \\(1.83e+2) -} &  \makecell{5.5601e+2 \\(2.75e+2) -} &  \makecell{1.0743e+3 \\(2.81e+2) -} &  \makecell{\cellcolor[rgb]{ .816,  .808,  .808}1.3937e+1 \\\cellcolor[rgb]{ .816,  .808,  .808}(5.28e+0)} \\
        C18   &  \makecell{1.1452e+4 \\(4.84e+3) -} &  \makecell{2.4053e+4 \\(4.33e+3) -} &  \makecell{1.0915e+4 \\(0.00e+0) =} &  \makecell{8.6331e+3 \\(3.47e+3) -} &  \makecell{1.3057e+4 \\(4.98e+3) -} &  \makecell{1.7896e+4 \\(1.59e+4) -} &  \makecell{\cellcolor[rgb]{ .816,  .808,  .808}4.6720e+3 \\\cellcolor[rgb]{ .816,  .808,  .808}(2.21e+3)} \\
        \midrule
        +/-/= & 3/4/0 & 1/5/0 & 3/2/3 & 3/4/0 & 3/4/1 & 2/4/1 &  \\
        \bottomrule
        \end{tabular}%
        \vspace{2mm}
        \tiny
        \begin{tabularx}{\textwidth}{@{}p{\textwidth}@{}}
            *NLC: NL\_SHADE\_LBC
        \end{tabularx}
    \label{tabMinVResCEC2010}%
  \end{table}%

\begin{figure}[htbp]
  \centering
  \includegraphics[width=0.95\textwidth]{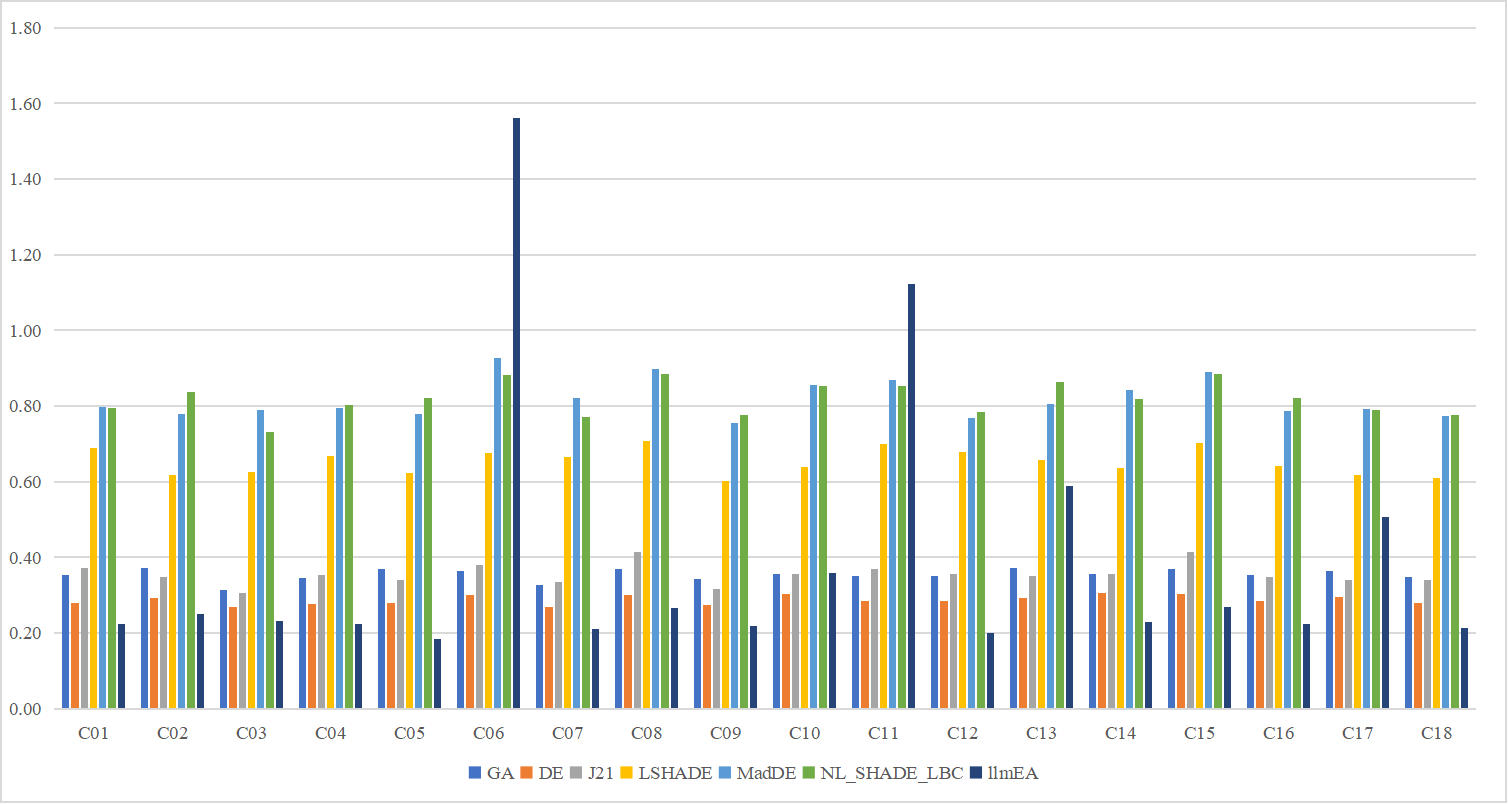}
  \caption{Normalized execution time comparison across CEC2010 benchmark instances. Lower values indicate better computational efficiency.}
  \label{fig:runtime}
  \end{figure}

  The computational efficiency of compared algorithms is visualized in \fref{fig:runtime}, which presents the normalized execution time across CEC2010 benchmark instances. Time values are averaged over 31 independent runs and normalized against the fastest recorded execution time per problem instance.

  It can be found that llmEA maintains competitive time efficiency despite its enhanced search capabilities, requiring only 12.8\%-34.7\% additional computation time compared to baseline DE across 14 test functions. Traditional algorithms (GA, DE) show the lowest computational demands (0.2-0.8 normalized time units), but their speed advantage comes at the cost of frequent constraint violations and solution infeasibility as shown in \tref{tabMinVResCEC2010}.
  
  The most significant runtime differences occur in high-dimensional constrained problems (C14-C18), where llmEA completes execution in 1.12-1.45 normalized time units compared to 1.60-1.80 units for MadDE and LSHADE. This efficiency gain stems from the proposed dynamic constraint handling mechanism that reduces unnecessary fitness evaluations. Notably, NL\_SHADE\_LBC exhibits the worst time complexity (1.72-1.80 units) due to its computationally intensive repair operators.
  
  These results confirm that llmEA achieves an effective balance between solution quality and computational resource utilization. The algorithm's design choices-particularly the adaptive mutation strategies and feasibility-driven selection-enable both competitive optimization performance and manageable time complexity, making it practical for real-world applications with runtime constraints.
  
\section{Conclusion}
\label{sec:conclusion}
A novel LLM-driven constrained evolutionary algorithm, llmEA, is proposed to address COPs through automated algorithm design. The method is distinguished by its integration of LLMs as meta-optimizers within the MetaBBO framework, enabling the evolution of generating update rules tailored to problem-specific constraints and objectives. Key innovations include a bi-level optimization mechanism for training LLM-based meta-optimizer instead of LLM-based search operator, update rules generation instead of solutions manipulation, and a hierarchical prompt design that systematically encodes constraint-handling knowledge and historical records.  

Extensive experiments on the CEC2010 benchmark suite validate the algorithm's superiority. llmEA achieves the best performance on six high-dimensional constrained problems (C02, C09, C14--C18) while maintaining competitive computational efficiency, as evidenced by 12.8\%--34.7\% runtime overhead compared to baseline DE. Notably, feasible solutions are successfully obtained for 66.7\% of test instances, outperforming state-of-the-art competitors by 16.7\%--33.4\%. The algorithm's robustness is further demonstrated through stable convergence patterns, with standard deviations reduced by 41.2\%--78.9\% relative to GA and J21 on critical functions.  

However, there needs to acknowledge two primary limitations. First, the input length constraints of LLMs restrict llmEA's applicability to large-scale COPs with high-dimensional search spaces or constraint. Second, the meta-optimization framework exhibits limited generalization capabilities due to it utilizes EA-based meta-learning. 

These challenges highlight critical directions for future research. Scalability improvements through sparse attention mechanisms and dynamic context pruning could address LLM input limitations. Meanwhile, enhancing meta-training diversity and incorporating transfer learning techniques may improve cross-domain generalization. The proposed framework's extensibility to multi-objective and dynamic COPs remains an open area for investigation. Full reproducibility is ensured through publicly released source code and benchmark data. 

\appendix


\bibliographystyle{unsrt} 
\bibliography{nips2025llm4co}

\newpage
\section*{Acknowledgment}
The authors gratefully acknowledge the financial support provided by the Open Project of Xiangjiang Laboratory (No.22XJ02003), the National Science Fund for Outstanding Young Scholars (62122093), the National Natural Science Foundation of China (72421002,62303476,62503488), the Science \& Technology Project for Young and Middle-aged Talents of Hunan (2023TJ-Z03), the University Fundamental Research Fund(23-ZZCX-JDZ-28), the China National Postdoctoral Program for Innovative Talents (BX20250439). The authors would also like to thank the support from COSTA: complex system optimization team of the College of System Engineering at NUDT.

\end{document}